# On the Entropy of Written Spanish

Fabio G. Guerrero, *Member, IEEE*

*Abstract*—This paper reports on results on the entropy of the Spanish language. They are based on an analysis of natural language for $n$-word symbols ($n$ = 1 to 18), trigrams, digrams, and characters. The results obtained in this work are based on the analysis of twelve different literary works in Spanish, as well as a 279917 word news file provided by the Spanish press agency EFE. Entropy values are calculated by a direct method using computer processing and the probability law of large numbers. Three samples of artificial Spanish language produced by a first-order model software source are also analyzed and compared with natural Spanish language.

*Index Terms*— Information theory, entropy rate, Spanish entropy, probability, stochastic processes.

## I. INTRODUCTION

Spanish is a language which is used by more than 400 million people in more than twenty countries, and which has been making its presence increasingly felt on the Internet [1]. Yet this language has not been as extensively researched at entropy level as some other languages. Accurate entropy calculations for the Spanish language are almost nonexistent, and the very few calculations which have been reported in the literature have usually been obtained by indirect methods. So it seems worth studying the average uncertainty content (i.e. entropy) for this language, with the aid of the computer processing capacity available at present. The aim of this paper is to report entropy values for Spanish found using a direct method based on calculating probability by counting symbols over long samples of text. The analysis reported in this paper was carried out using a software program written in Matlab 7.0 called IT-TUTOR-UV [2]. Twelve literary works of both ibero-american and other countries' literature available in Spanish were analyzed. Also, a large archive file of news provided to the author by the Spanish press agency EFE was included. Additionally three samples of artificial language produced by a first-order source software model available at IT-TUTOR-UV were analyzed, and the results were compared with results from the natural Spanish samples.

Several approaches have been devised for finding the entropy of a language. C. E. Shannon initially showed in [3] that one way to calculate the entropy of a language is through the probability of longer and longer sequences, adding over all sequences $B_i$ of $N$ symbols to find

$$G_N = \frac{1}{N} \sum_i p(B_i) \log_2 p(B_i) \qquad (1)$$

and then taking the limit

$$H = \lim_{N \to \infty} G_N \qquad (2)$$

One unfavorable aspect of finding $H$ by using direct methods such as the one suggested by (2) is that calculating $p(B_i)$ can become an extremely complex computing problem as $N$ increases. As is well known, for a sequence of $n$ symbols the number of typical sequences will be close to $2^{nH(X)}$, where $H(X)$ is the source entropy and $n$ is the length of the sequence. For example, assuming an entropy $H(X)$ of 1 bit/character, for a text of 900 letters there will be around $8.45 \times 10^{270}$ typical sequences. Hence to get some reasonable reliable statistical values for finding entropy using this method it would be necessary to count over an extremely large search space. However, for natural languages, dependency between symbols may tend to disappear after a reasonably large number of characters or words. Hence, it might not be necessary to obtain values of $p(B_i)$ for extremely large values of $n$ before (2) converges to the limit.

C. E. Shannon also presented in [3] and then in [4] an alternative way to calculate the entropy $H$ of a natural language by means of a series of approximations $F_0$, $F_1$, $F_2$, ... given by

$$F_N = -\sum_{i,j} p(B_i, j) \log_2 p(j \mid B_i) = -\sum_{i,j} p(B_i, j) \log_2 p(j, B_i) + \sum_i p(B_i) \log_2 p(B_i) \qquad (3)$$

where $B_i$ is a block of $N$-1 symbols, $j$ is the symbol which appears after $B_i$, and $p(B_i, j)$ is the probability of the $N$-symbol ($B_i$, $j$), and $p(j \mid B_i)$ is the conditional probability of symbol $j$ after block $B_i$. In this approach, longer sequences are progressively taken and used to calculate conditional entropy values, and then $H$ is calculated by the limit:

Fabio G. Guerrero, is with Universidad del Valle, Cali, COLOMBIA (South America) (phone: +57.3392140 ext. 109; e-mail: fguerrer@ univalle.edu.co).

$$H = \lim_{N \to \infty} F_N \qquad (4)$$

In (4), $F_N$, in bit/symbol, is the entropy of an $N$-symbol which measures the amount of information due to the statistics considering $N$ as consecutive symbols of text.

In [4] C. E. Shannon used a human prediction approach for finding the entropy of English values, finding bounds for printed English of between 0.6 and 1.3 bits/letter considering 100-letter sequences. Cover and King [5] estimated an entropy value of 1.25 bits per character for English using gambling estimations.

Another opportunity for calculating the entropy rate of a language is to use ideal source coders since, by definition, this kind of coder should compress to the entropy limit. In [6], for instance, a value of 1.46 bits per character was reported for entropy of English by means of data compression. In [7], universal data compression algorithms were used to estimate the entropy of the fruit fly genetic code. No matter what the source coding method, coding close to the entropy rate of English continues to be a challenging subject [8].

As for the Spanish language, values of 4.70, 4.015, and 1.97 bits/letter for $F_0$, $F_1$ and $F_W$ respectively were reported in [9] using an extrapolation technique on frequency data obtained from a sample of 6513 different words.

Communication problems have, classically, been portrayed on three levels where Level A deals with how accurately the symbols that a source of information produces can be transmitted; Level B with how accurately the transmitted symbols produce the desired meaning, and Level C with the extent to which the meaning given to the message by the receiver produces the desired action the message was intended to produce. From this perspective entropy calculation would be at the lowest level of language analysis, i.e. level A, because it only takes into account source symbol statistics and their statistical dependence, without any further consideration of more intelligent aspects of language such as grammar, semantics, and punctuation marks which can considerably change the meaning of a sentence, and so on.

This paper is organized as follows: in section II the methodology used to obtain all the values reported in this paper is discussed; in section III the results of the observations are presented; section IV presents a discussion of the most relevant results and, finally, in section V the conclusions of this work are summarized. Aspects such as the analysis of grammar, semantics and similar aspects of Spanish, as well as computational complexity and compression theory, are beyond the scope of this paper. Support material for this work is available at [10].

## II. Methodology

The twelve modern twentieth and twenty first century novels used in this paper as samples of literary Spanish were obtained from public libraries available on the Internet such as librodot [11] and the virtual library Miguel de Cervantes [12]. The selection of the literary works was done without any particular consideration of publication period, author's country of origin, and suchlike. As a sample of the Spanish used in the news, a news archive of 279917 words provided by the Spanish press agency EFE, one of the most important press agencies in Spain, was used. Finally three samples of text generated by the IT-TUTOR-UV artificial Spanish source, which uses a first-order model (i.e. words with the correct statistics for Spanish but which considers words as statistically independent symbols), were analyzed. All the selected material was processed using a generic personal computer with 1GB RAM and a 2-GHz double-core CPU.

The twelve literary works (LW) chosen were the following:

LW1 = *Amalia* by José Mármol

LW2 = Cien Años de Soledad (*One Hundred Years of Solitude*) by Gabriel García Márquez

LW3 = Al primer vuelo (*At the First Flight*) by José María de Pereda

LW4 = Harry Potter y la Cámara Secreta (*Harry Potter and the Secret Chamber*) by J.K. Rowling

LW5 = *Maria* by Jorge Isaacs

LW6 = Colmillo Blanco (*White Fang*) by Jack London

LW7 = El archipiélago en llamas (*The Archipelago in Flames*) by Jules Verne

LW8 = El Cisne de Vilamorta (*The Swan of Vilamorta*) by Emilia Pardo Bazán

LW9 = *Tristana* by Benito Pérez Galdós

LW10 = Cuarto Menguante (*Waning Gibbous Moon*) by Enrique Cerdán Tato

LW11 = Historia de la Vida del Buscón (*The Scavenger*) by Francisco de Quevedo

LW12 = Creció Espesa la Yerba (*Thick Grew the Grass*) by Carmen Conde

The three samples of artificial Spanish text (without punctuation marks) produced by the IT-TUTOR-UV were named AT1, AT2, and AT3, having the same number of words as LW1, LW2 and LW3 respectively for comparison purposes.

In the text generating mode, the IT-TUTOR-UV employs a database of the 81323 most frequent words as compiled by Alameda & Cuetos from a corpus of 1950375 words of written Spanish [13]. The twenty-second version of the Dictionary of the Royal Academy of the Spanish Language (DRAS) has 88431 lemmas (entries) with 161962 definitions (i.e. meanings for the words according to the context in which they appear). Therefore, the number of words compiled by the Alameda & Cuetos database is quite close to the number of words in the DRAS.

Table I shows the length, in words, of each sample of natural Spanish analyzed in this paper. The parameter $\alpha$, average word length is given by $\sum L_i p_i$, where $L_i$ is the length in characters of the $i$-th word, and $p_i$ the probability of the $i$-th word.

TABLE I
AVERAGE WORD LENGTH FOR SEVERAL NATURAL SPANISH TEXTS

| Work | Number of words | Different words | $\alpha$ |
|---|---|---|---|
| EFE | 279917 | 27782 | 4.80 |
| LW1 | 231860 | 18874 | 4.51 |
| LW2 | 137783 | 15970 | 4.73 |
| LW3 | 100797 | 13163 | 4.35 |
| LW4 | 91388 | 10882 | 4.60 |
| LW5 | 88376 | 12680 | 4.45 |
| LW6 | 81223 | 10027 | 4.58 |
| LW7 | 61386 | 8470 | 4.73 |
| LW8 | 53035 | 11857 | 4.65 |
| LW9 | 52571 | 10580 | 4.48 |
| LW10 | 49835 | 12945 | 4.95 |
| LW11 | 42956 | 7660 | 4.23 |
| LW12 | 27813 | 6087 | 4.48 |

In Table I the weighted average value of $\alpha$ is 4.61 letters per word, and the sum total of the second column (number of words) is 1298940.

Table II shows the length, in words, of the three artificial text samples generated using the IT-TUTOR-UV.

| Text | Length (words) | Different words | α |
|------|------|------|------|
| AT1 | 231860 | 28344 | 4.70 |
| AT2 | 137783 | 21129 | 4.69 |
| AT3 | 100797 | 17530 | 4.70 |

In Table II the average value of α is 4.70 letters per word, very close to the value of α calculated over the entire Alameda & Cuetos database of 4.6978 letters per word. The sum of words (second column) in Table II is 470440.

To calculate the different entropy values, the frequency of each symbol is obtained and then the value of its probability: $p(B_i) \approx n_{Bi}/n_{total}$, is used in the classic entropy formula $H = -\sum_i p_i \log_2 p_i$ .

*N*-symbol entropy for values of *N* equal to 1, 2 and 3 were calculated first, obtaining entropy values for characters, digrams, and trigrams. After trigrams, we started considering words instead of letters, for two reasons: firstly, that there are many trigrams which are at the same time words; and secondly and mainly because the constituent elements of the Spanish language are words. Hence, entropy values for 1-word, 2-word, 3-word, up to 18-word symbols were calculated. The eighteen-word value was found to be long enough to guarantee having equiprobable symbols (*n*-words) for all cases except for the EFE archive. In the archive some partial news is repeated as part of a larger updated news report. When finding the frequency for *n*-word symbols the assumption that the text under analysis was produced by a source which produces statistically independent *n*-word symbols is implicitly being made. Values of *n* for which the maximum value of entropy was produced were identified, as well as values of *n* from which all symbols present in the text are equiprobable, i.e. none of them repeat more than once.

## III. RESULTS

### A. Entropy Values

Table III shows the values of entropy for characters (considering both uppercase and lowercase alphanumeric characters, spaces, and punctuation marks), digrams (considering only alphanumeric uppercase and lowercase characters), and trigrams (considering only alphanumeric uppercase and lowercase characters) for literary works LW1 to LW12 and the EFE archive. For both digrams and trigrams, those symbols bridging two words were taken into account; for instance, the trigrams DOD and DIA in the Spanish words LINDO DIA (beautiful day).

TABLE III
CHARACTER, DIGRAM, AND TRIGRAM ENTROPY VALUES FOR NATURAL SPANISH

| Work | $H_{char}$ | $H_{digram}$ | $H_{trigram}$ |
|------|------|------|------|
| EFE | 4.52 | 8.20 | 11.35 |
| LW1 | 4.38 | 7.92 | 11.04 |
| LW2 | 4.28 | 7.78 | 10.83 |
| LW3 | 4.36 | 7.84 | 10.92 |
| LW4 | 4.48 | 8.08 | 11.14 |
| LW5 | 4.39 | 7.89 | 11.00 |
| LW6 | 4.32 | 7.85 | 10.86 |
| LW7 | 4.39 | 7.88 | 10.86 |
| LW8 | 4.38 | 7.90 | 11.03 |
| LW9 | 4.37 | 7.89 | 11.01 |
| LW10 | 4.33 | 7.89 | 11.03 |
| LW11 | 4.34 | 7.84 | 10.86 |
| LW12 | 4.41 | 7.90 | 10.91 |

In Table III the weighted average values for $H_{char}$, $H_{digram}$ and $H_{trigram}$ are 4.40 bits/character, 7.96 bits/digram, and 11.05 bits/trigram respectively.

Table IV shows the values of entropy for characters, digrams, and trigrams for AT2, AT2, and AT3.

TABLE IV
CHARACTER, DIGRAM AND TRIGRAM ENTROPY VALUES FOR ARTIFICIAL SAMPLES
OF SPANISH TEXT

| Text | $H_{char}$ | $H_{digram}$ | $H_{trigram}$ |
|------|------|------|------|
| AT1 | 4.09 | 7.65 | 10.79 |
| AT2 | 4.09 | 7.65 | 10.78 |
| AT3 | 4.09 | 7.64 | 10.77 |

In Table IV the weighted average for $H_{char}$, $H_{digram}$ and $H_{trigram}$ are 4.09 bits/character, 7.65 bits/digram, and 10.78 bits/trigram respectively. The differences from the average values in Table III can be explained because the Alameda and Cuetos database employs only lowercase letters.

Table V shows the values of entropy for $n$-word symbols from $n = 1$ to 18 for literary works LW1 to LW12 and the EFE archive. The values in italics indicate that the symbols found were equiprobable. The case of equiprobable symbols for the EFE archive occurred at a higher value of $n = 38$ words, as expected, because the language of the news reuses news as part of larger updated reports.

At the literary level, it can be observed from Table I that the authors of the literary works analyzed in this work tend to use a relatively small set of words, including Nobel Prize winning writer Gabriel García Márquez. The most prolific author only used nearly 21% of the lemmas defined by the DRAS. It can also be observed that, in general, the greater the number of words in the literary work, the larger the number of different words used.

TABLE V
ENTROPY VALUES $H_{N\text{-WORD}}$ FOR LW1 TO LW12 AND EFE ARCHIVE

| $H_{n\text{-word}}$ | EFE | LW1 | LW2 | LW3 | LW4 | LW5 | LW6 | LW7 | LW8 | LW9 | LW10 | LW11 | LW12 | Weighted average |
|---|---|---|---|---|---|---|---|---|---|---|---|---|---|---|
| 1 | 10.43 | 9.91 | 9.80 | 9.71 | 9.86 | 9.97 | 9.70 | 9.69 | 10.19 | 10.00 | 10.32 | 9.51 | 9.77 | 9.99 |
| 2 | 15.22 | 14.91 | 14.35 | 14.26 | 14.10 | 14.21 | 13.97 | 13.60 | 13.92 | 13.86 | 13.82 | 13.43 | 13.10 | 14.43 |
| 3 | 16.19 | 15.98 | 15.24 | 14.91 | 14.72 | 14.73 | 14.57 | 14.12 | 14.06 | 14.04 | 13.97 | 13.72 | 13.13 | 15.14 |
| 4 | 16.03 | 15.79 | 15.02 | 14.61 | 14.45 | 14.42 | 14.29 | 13.88 | 13.69 | 13.68 | 13.59 | 13.39 | 12.76 | 14.90 |
| 5 | 15.75 | 15.49 | 14.74 | 14.30 | 14.15 | 14.11 | 13.99 | 13.58 | 13.37 | 13.36 | 13.28 | 13.07 | 12.44 | 14.60 |
| 6 | 15.50 | 15.24 | 14.48 | 14.04 | 13.89 | 13.85 | 13.72 | 13.32 | 13.11 | 13.10 | 13.02 | 12.81 | 12.18 | 14.34 |
| 7 | 15.28 | 15.02 | 14.26 | 13.81 | 13.67 | 13.62 | 13.50 | 13.10 | 12.89 | 12.87 | 12.80 | 12.58 | 11.96 | 14.12 |
| 8 | 15.09 | 14.82 | 14.07 | 13.62 | 13.48 | 13.43 | 13.31 | 12.91 | 12.69 | 12.68 | 12.60 | 12.39 | 11.76 | 13.93 |
| 9 | 14.92 | 14.65 | 13.90 | 13.45 | 13.31 | 13.26 | 13.14 | 12.74 | 12.52 | 12.51 | 12.43 | 12.22 | 11.59 | 13.76 |
| 10 | 14.77 | 14.50 | 13.75 | 13.30 | 13.16 | 13.11 | 12.99 | 12.58 | 12.37 | 12.36 | 12.28 | 12.07 | 11.44 | 13.61 |
| 11 | 14.63 | 14.36 | 13.61 | 13.16 | 13.02 | 12.97 | 12.85 | 12.45 | 12.24 | 12.22 | 12.15 | 11.93 | 11.30 | 13.47 |
| 12 | 14.51 | 14.24 | 13.49 | 13.04 | 12.89 | 12.85 | 12.72 | 12.32 | 12.11 | 12.10 | 12.02 | 11.81 | 11.18 | 13.34 |
| 13 | 14.39 | 14.12 | 13.37 | 12.92 | 12.78 | 12.73 | 12.61 | 12.21 | 11.99 | 11.98 | 11.90 | 11.69 | 11.06 | 13.23 |
| 14 | 14.29 | 14.02 | 13.26 | 12.81 | 12.67 | 12.62 | 12.50 | 12.10 | 11.89 | 11.87 | 11.80 | 11.58 | 10.95 | 13.12 |
| 15 | 14.19 | 13.92 | 13.17 | 12.71 | 12.57 | 12.52 | 12.40 | 12.00 | 11.79 | 11.77 | 11.70 | 11.48 | 10.86 | 13.02 |
| 16 | 14.09 | 13.82 | 13.07 | 12.62 | 12.48 | 12.43 | 12.31 | 11.91 | 11.69 | 11.68 | 11.60 | 11.39 | 10.76 | 12.93 |
| 17 | 14.01 | 13.74 | 12.98 | 12.53 | 12.39 | 12.34 | 12.22 | 11.82 | 11.61 | 11.59 | 11.52 | 11.30 | 10.68 | 12.84 |
| 18 | 13.92 | 13.65 | 12.90 | 12.45 | 12.31 | 12.26 | 12.14 | 11.74 | 11.52 | 11.51 | 11.43 | 11.22 | 10.59 | 12.76 |

Fig. 1 shows the entropy versus $n$-word symbol curve for EFE, LW12, LW6, LW2, and LW1. The rest of literary works exhibited the same curve shapes with values in between, as can be easily observed in Table V.

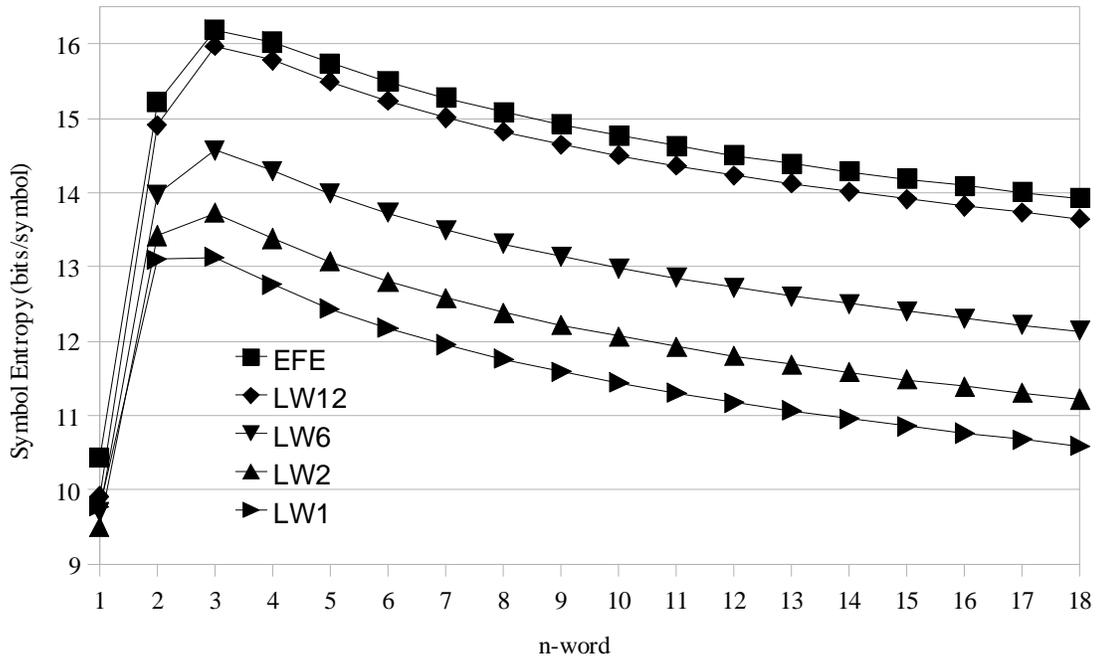

Fig. 1. Symbol entropy versus *n*-word symbol curve for EFE, LW12, LW6, LW2 and LW1.

Table VI shows the entropy values for *n*-word symbols from *n* = 1 to 18 for AT1, AT2, and AT3.

TABLE VI
ENTROPY VALUES $H_{n\text{-word}}$ FOR ARTIFICIAL SAMPLES OF
SPANISH TEXT

| $H_{n\text{-word}}$ | AT1 | AT2 | AT3 |
|---|---|---|---|
| 1 | 10.22 | 10.13 | 10.09 |
| 2 | 15.54 | 14.99 | 14.68 |
| 3 | 16.17 | 15.44 | 15.00 |
| 4 | 15.82 | 15.07 | 14.62 |
| 5 | 15.50 | 14.75 | 14.30 |
| 6 | 15.24 | 14.49 | 14.04 |
| 7 | 15.02 | 14.26 | 13.81 |
| 8 | 14.82 | 14.07 | 13.62 |
| 9 | 14.65 | 13.90 | 13.45 |
| 10 | 14.50 | 13.75 | 13.30 |
| 11 | 14.36 | 13.61 | 13.16 |
| 12 | 14.24 | 13.49 | 13.04 |
| 13 | 14.12 | 13.37 | 12.92 |
| 14 | 14.02 | 13.26 | 12.81 |
| 15 | 13.92 | 13.17 | 12.71 |
| 16 | 13.82 | 13.07 | 12.62 |
| 17 | 13.74 | 12.98 | 12.53 |
| 18 | 13.65 | 12.90 | 12.45 |

Fig. 2 shows the curves of symbol entropy versus *n*-word symbols for AT1, AT2, and AT3.

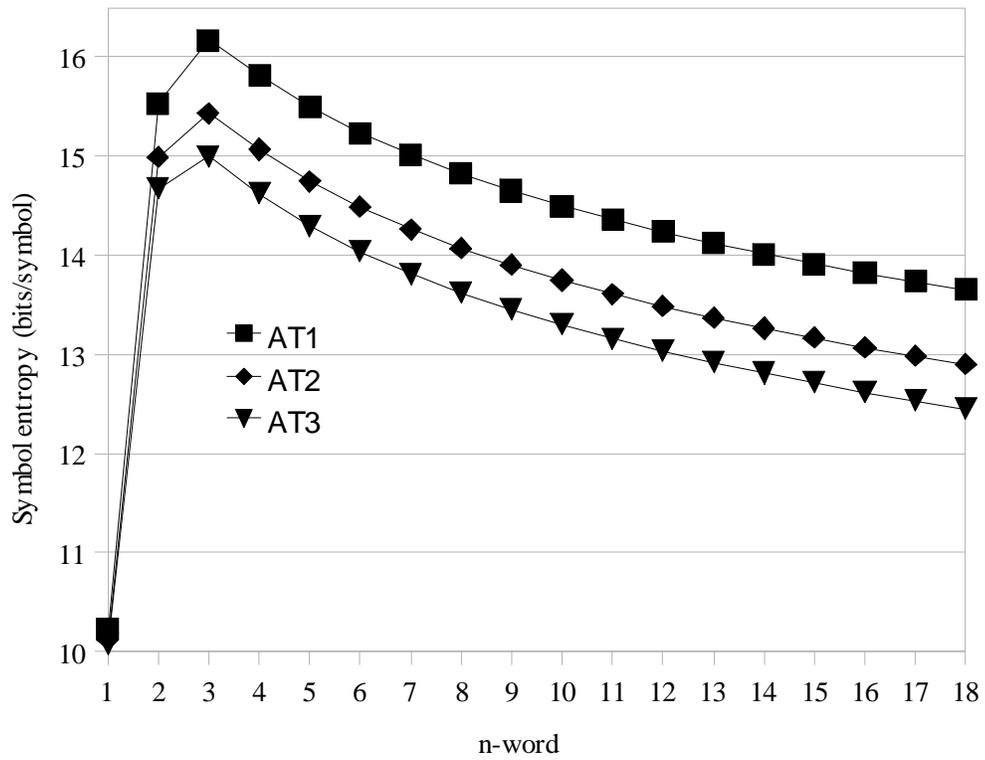

Fig. 2.  Symbol entropy versus n-word symbol curves for AT1, AT2 and AT3.

Table VII shows the entropy in bits/character given by:

$$H_{char} = \frac{H_{symbol}}{k \cdot \alpha} \qquad (5)$$

where $k$ is number of words and $\alpha$ is the average word length.

TABLE VII
ENTROPY VALUES FOR NATURAL SPANISH TEXT IN BITS/CHARACTER

| $H_{char}$ | EFE | LW1 | LW2 | LW3 | LW4 | LW5 | LW6 | LW7 | LW8 | LW9 | LW10 | LW11 | LW12 | Weighted average |
|---|---|---|---|---|---|---|---|---|---|---|---|---|---|---|
| 1 | 2.17 | 2.20 | 2.07 | 2.23 | 2.15 | 2.24 | 2.12 | 2.05 | 2.19 | 2.23 | 2.09 | 2.25 | 2.18 | 2.17 |
| 2 | 1.58 | 1.65 | 1.52 | 1.64 | 1.53 | 1.60 | 1.53 | 1.44 | 1.50 | 1.55 | 1.40 | 1.59 | 1.46 | 1.57 |
| 3 | 1.12 | 1.18 | 1.07 | 1.14 | 1.07 | 1.10 | 1.06 | 1.00 | 1.01 | 1.05 | 0.94 | 1.08 | 0.98 | 1.10 |
| 4 | 0.83 | 0.87 | 0.79 | 0.84 | 0.79 | 0.81 | 0.78 | 0.73 | 0.74 | 0.76 | 0.69 | 0.79 | 0.71 | 0.81 |
| 5 | 0.66 | 0.69 | 0.62 | 0.66 | 0.62 | 0.63 | 0.61 | 0.57 | 0.58 | 0.60 | 0.54 | 0.62 | 0.56 | 0.63 |
| 6 | 0.54 | 0.56 | 0.51 | 0.54 | 0.50 | 0.52 | 0.50 | 0.47 | 0.47 | 0.49 | 0.44 | 0.50 | 0.45 | 0.52 |
| 7 | 0.45 | 0.48 | 0.43 | 0.45 | 0.42 | 0.44 | 0.42 | 0.40 | 0.40 | 0.41 | 0.37 | 0.43 | 0.38 | 0.44 |
| 8 | 0.39 | 0.41 | 0.37 | 0.39 | 0.37 | 0.38 | 0.36 | 0.34 | 0.34 | 0.35 | 0.32 | 0.37 | 0.33 | 0.38 |
| 9 | 0.35 | 0.36 | 0.33 | 0.34 | 0.32 | 0.33 | 0.32 | 0.30 | 0.30 | 0.31 | 0.28 | 0.32 | 0.29 | 0.33 |
| 10 | 0.31 | 0.32 | 0.29 | 0.31 | 0.29 | 0.29 | 0.28 | 0.27 | 0.27 | 0.28 | 0.25 | 0.29 | 0.26 | 0.30 |
| 11 | 0.28 | 0.29 | 0.26 | 0.28 | 0.26 | 0.27 | 0.26 | 0.24 | 0.24 | 0.25 | 0.22 | 0.26 | 0.23 | 0.27 |
| 12 | 0.25 | 0.26 | 0.24 | 0.25 | 0.23 | 0.24 | 0.23 | 0.22 | 0.22 | 0.23 | 0.20 | 0.23 | 0.21 | 0.24 |
| 13 | 0.23 | 0.24 | 0.22 | 0.23 | 0.21 | 0.22 | 0.21 | 0.20 | 0.20 | 0.21 | 0.19 | 0.21 | 0.19 | 0.22 |
| 14 | 0.21 | 0.22 | 0.20 | 0.21 | 0.20 | 0.20 | 0.19 | 0.18 | 0.18 | 0.19 | 0.17 | 0.20 | 0.17 | 0.20 |
| 15 | 0.20 | 0.21 | 0.19 | 0.19 | 0.18 | 0.19 | 0.18 | 0.17 | 0.17 | 0.18 | 0.16 | 0.18 | 0.16 | 0.19 |
| 16 | 0.18 | 0.19 | 0.17 | 0.18 | 0.17 | 0.17 | 0.17 | 0.16 | 0.16 | 0.16 | 0.15 | 0.17 | 0.15 | 0.18 |
| 17 | 0.17 | 0.18 | 0.16 | 0.17 | 0.16 | 0.16 | 0.16 | 0.15 | 0.15 | 0.15 | 0.14 | 0.16 | 0.14 | 0.16 |
| 18 | 0.16 | 0.17 | 0.15 | 0.16 | 0.15 | 0.15 | 0.15 | 0.14 | 0.14 | 0.14 | 0.13 | 0.15 | 0.13 | 0.15 |

Fig. 3 shows the entropy versus $n$-word symbol curve for LW1. It should be observed that in Fig. 3, if $n$ tends to infinite then the entropy $H$ would tend to zero because the lengths of the text are finite.

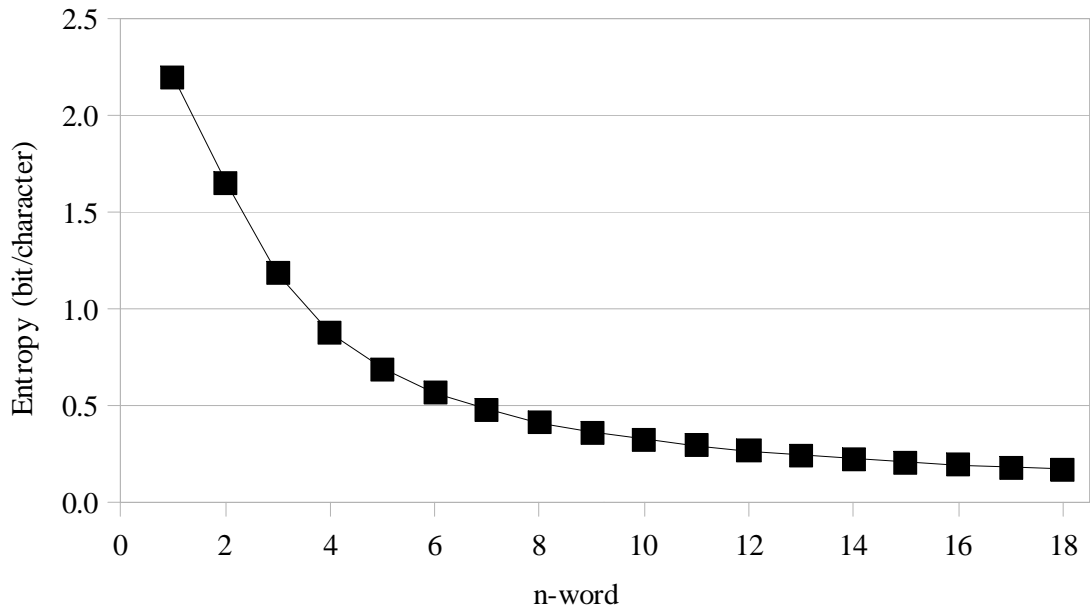

Fig. 3.  $H$ (bit/character) versus $n$-word symbol curve for LW1.

*B.  Symbol Entropy Processing Time*

Fig. 4 shows the time required for the calculation of character, digram, and trigram entropy for the EFE archive.

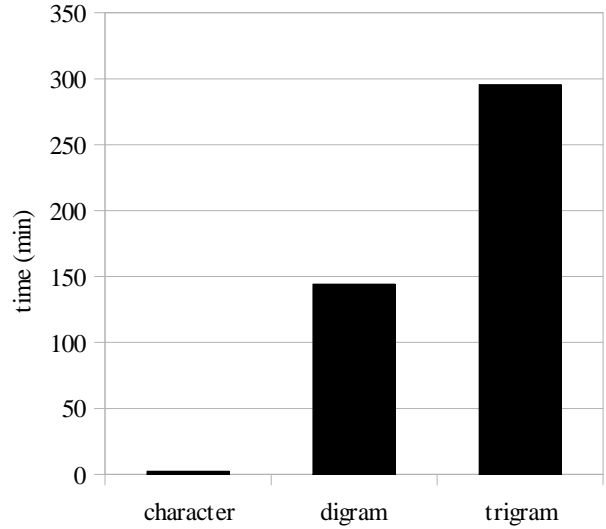

Fig. 4. Approximate computing time for characters, digrams, and trigrams (in minutes) for EFE archive.

Fig. 5 shows the time required for the calculation of *n*-word entropy for the EFE archive. For the rest of the works analyzed, the shapes of both Figures 4 and 5 repeated similarly with different scale values.

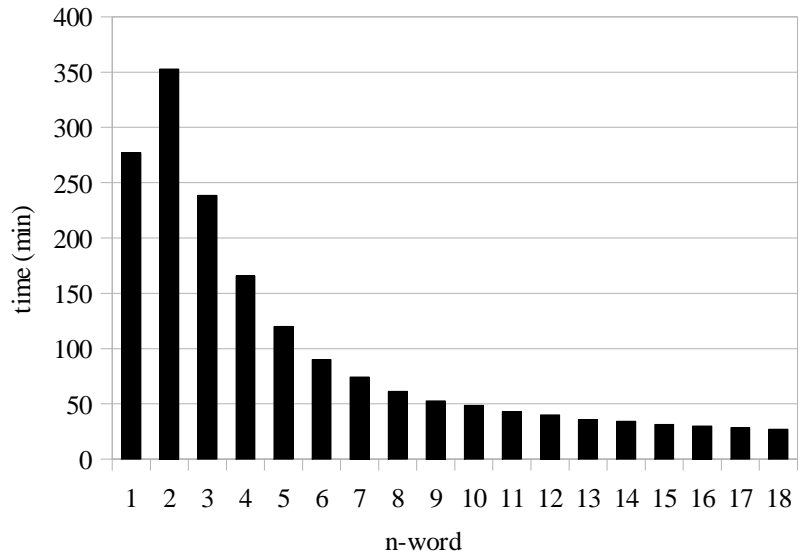

Fig. 5. Approximate computing time for n-words (minutes) for EFE archive.

*C. Log-log Plots from Frequency Analysis*

Although literary works LW1 to LW12 were analyzed counting symbols up to a length of eighteen words, and for the EFE

news archive up to 40-word symbols, in this section only log-log plots for 3-word, 2-word, 1-word, trigram, digram, and single characters for the EFE archive are presented.

Fig. 6 shows the log-log plot for 1-word, 2-word, and 3-word symbols for the EFE archive. The EFE archive contained 82656 different 3-word symbols, 79704 different 2-word symbols, and 27782 different 1-word symbols. Log-log plots for the literary works LW1 to LW12 were found to be quite similar to those of Fig. 6 for 2-word and 1-word symbols. For 3-word symbols there was a little more discrepancy because a lot more text may need to be analyzed in order to get accurate 3-word probability values.

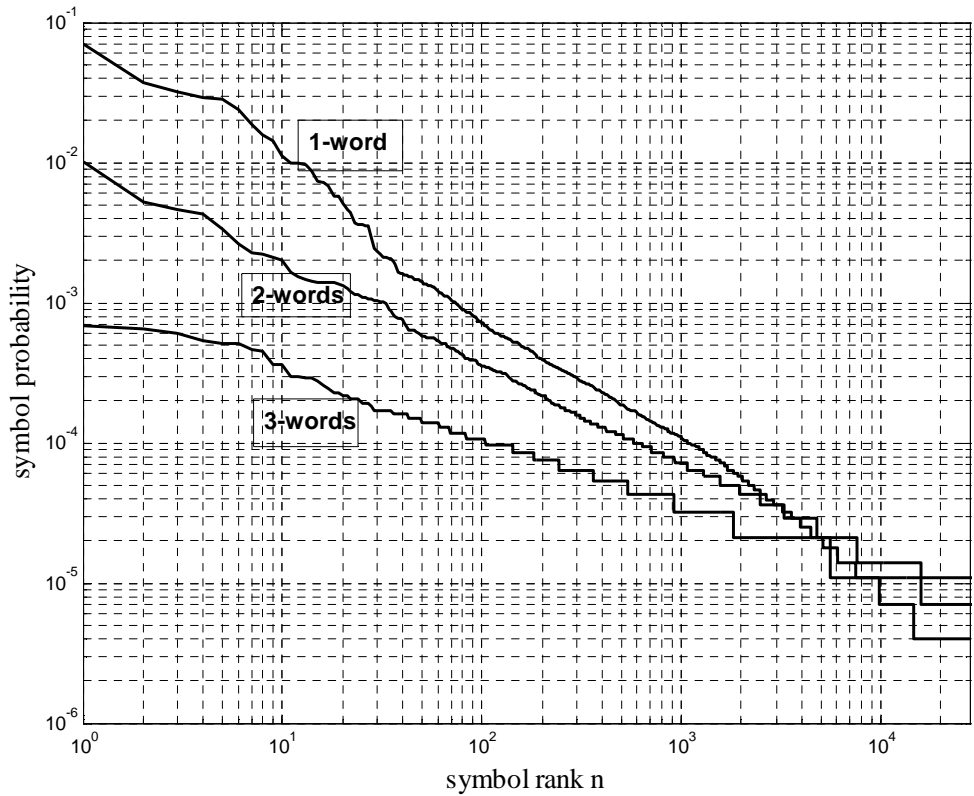

Fig. 6. Symbol rank versus symbol probability for EFE archive.

Fig. 7 shows the log-log plot for trigrams, digrams, and characters for the EFE archive, which contained 19208 different trigrams, 3046 different digrams, and 111 different characters. Similarly, log-log plots for the literary works were found to be close to those of Fig. 7 for trigrams, digrams and characters, as expected.

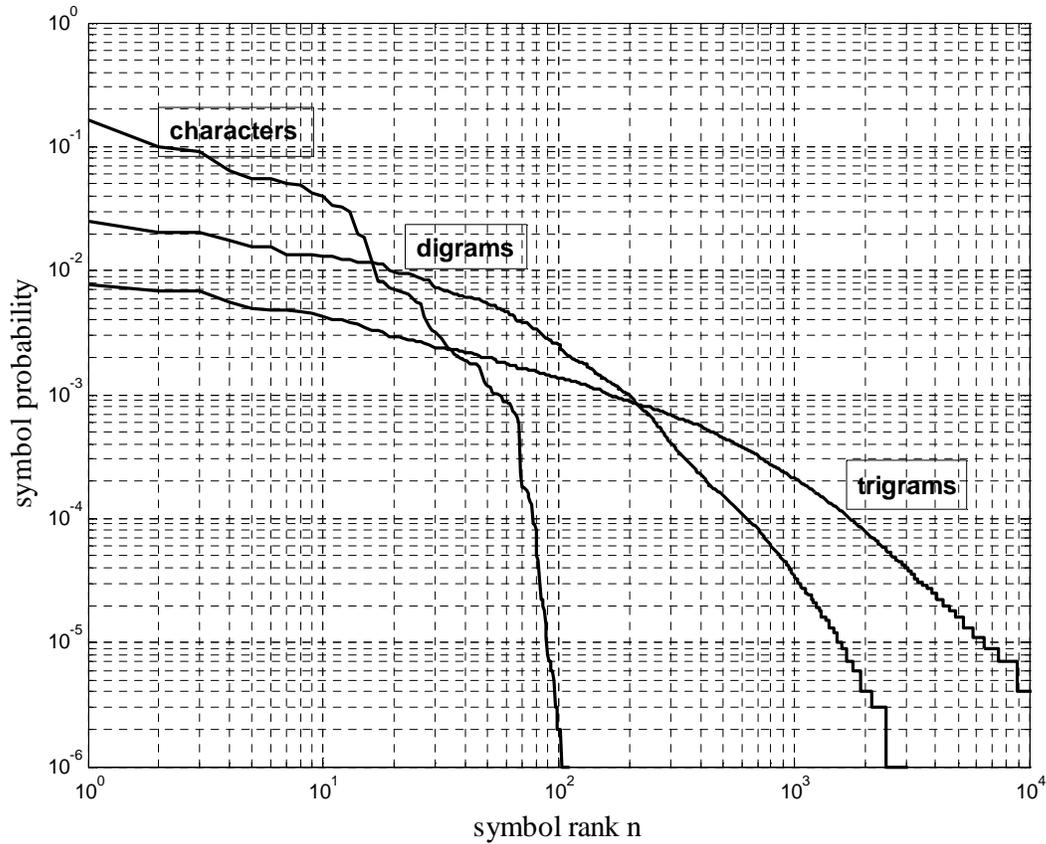

Fig. 7.  *n*-gram rank versus *n*-gram probability for EFE archive.

## IV.  DISCUSSION

To start this section it may be interesting to have a look at the $p_i \log_2 p_i$ function first. Fig. 8 shows a plot of the function $p_i \log_2 p_i$. This function has its maximum value, 0.530738, at $p_i = 0.36788$. This means any value of $p_i$ less than 0.36788 will be making a directly proportional contribution to the entropy $H$.

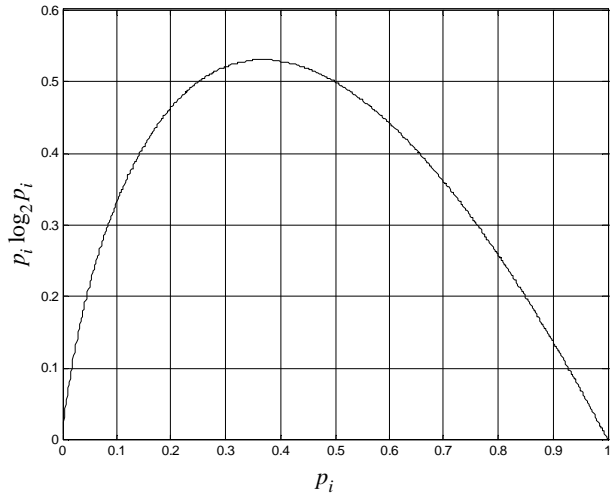

Fig. 8. Plot of the $p_i \log_2 p_i$ function.

### A. Entropy calculation from the statistics of Spanish

Ignoring spaces and punctuation marks the Spanish alphabet can be considered an alphabet of 33 symbols, hence by definition $F_0$ equals $\log_2 33$ or 5.0444 bits per letter. Including the digits 0 through 9 the value of $F_0$ would be $\log_2 42$ or 5.3923; including the distinction between lowercase and uppercase characters $\log_2 75$ etc. The value of $F_1$ depends on the frequency of single characters and is given by $F_1 = -\sum_i p(i) \log_2 p(i)$ in bits per character.

Using (3) the approximation for digrams $F_2$ is given by

$$F_2 = -\sum_{i,j} p(i,j) \log_2 p(j \mid i) = -\sum_{i,j} p(i,j) \log_2 p(i,j) + \sum_i p(i) \log_2 p(i) \qquad (6)$$

Similarly the entropy for trigrams $F_3$ is given by:

$$F_3 = -\sum_{i,j,k} p(i,j,k) \log_2 p(k \mid ij) = -\sum_{i,j,k} p(i,j,k) \log_2 p(i,j,k) + \sum_{i,j} p(i,j) \log_2 p(i,j) \qquad (7)$$

Tables VIII and IX show the values of $F_2$ and $F_3$ for the works considered in this paper according to (7).

TABLE VIII
ENTROPY FOR $F_2$ AND $F_3$ NATURAL SPANISH

| Work | $F_2$ | $F_3$ |
|------|-------|-------|
| EFE | 3.68 | 3.15 |
| LW1 | 3.53 | 3.12 |
| LW2 | 3.50 | 3.05 |
| LW3 | 3.48 | 3.08 |
| LW4 | 3.60 | 3.05 |
| LW5 | 3.51 | 3.10 |
| LW6 | 3.53 | 3.01 |
| LW7 | 3.50 | 2.98 |
| LW8 | 3.52 | 3.13 |
| LW9 | 3.52 | 3.12 |
| LW10 | 3.56 | 3.14 |
| LW11 | 3.50 | 3.02 |
| LW12 | 3.49 | 3.01 |

In Table VIII the weighted average values for $F_2$ and $F_3$ are 3.56 bits/character and 3.09 bits/character respectively.

TABLE IX
CHARACTER, DIGRAM AND TRIGRAM ENTROPY VALUES FOR ARTIFICIAL SAMPLES
OF SPANISH TEXT

| Text | $F_2$ | $F_3$ |
|------|-------|-------|
| AT1 | 3.56 | 3.14 |
| AT2 | 3.55 | 3.13 |
| AT3 | 3.55 | 3.13 |

In Tables III to and IV if word bridging trigrams or digrams had not been taken into consideration the entropy would have been slightly lower.

Also when $n$ is sufficiently large, the space symbol will be almost completely redundant, i.e. its probability will be very high, so that its contribution to uncertainty (entropy) will be low, producing a slight reduction of the source entropy. The entropy per character considering $k$-word symbols, including the space symbol, can then be approximated by:

$$H_{char} \approx \frac{H_{symbol}}{k \cdot (\alpha + 1)} \qquad (8)$$

where $\alpha$ is the average word length of the language. Table X shows the entropy (bits/character) obtained including the space symbol, using the approximation given by (8).

TABLE X
APPROXIMATED ENTROPY (BITS/CHARACTER) INCLUDING THE SPACE SYMBOL

| $H_{char}$ | EFE | LW1 | LW2 | LW3 | LW4 | LW5 | LW6 | LW7 | LW8 | LW9 | LW10 | LW11 | LW12 | Weighted Average |
|---|---|---|---|---|---|---|---|---|---|---|---|---|---|---|
| 1 | 2.171 | 2.196 | 2.070 | 2.233 | 2.145 | 2.241 | 2.118 | 2.049 | 2.193 | 2.235 | 2.087 | 2.249 | 2.183 | 2.166 |
| 2 | 1.311 | 1.352 | 1.252 | 1.333 | 1.260 | 1.304 | 1.252 | 1.188 | 1.233 | 1.265 | 1.162 | 1.284 | 1.196 | 1.286 |
| 3 | 0.930 | 0.966 | 0.886 | 0.929 | 0.877 | 0.901 | 0.871 | 0.822 | 0.830 | 0.855 | 0.783 | 0.875 | 0.800 | 0.900 |
| 4 | 0.690 | 0.716 | 0.655 | 0.683 | 0.645 | 0.662 | 0.640 | 0.606 | 0.606 | 0.624 | 0.572 | 0.640 | 0.583 | 0.664 |
| 5 | 0.543 | 0.562 | 0.514 | 0.535 | 0.506 | 0.518 | 0.501 | 0.474 | 0.474 | 0.488 | 0.447 | 0.500 | 0.454 | 0.520 |
| 6 | 0.445 | 0.461 | 0.421 | 0.437 | 0.414 | 0.424 | 0.410 | 0.388 | 0.387 | 0.399 | 0.365 | 0.408 | 0.371 | 0.426 |
| 7 | 0.376 | 0.389 | 0.355 | 0.369 | 0.349 | 0.357 | 0.346 | 0.327 | 0.326 | 0.336 | 0.307 | 0.344 | 0.312 | 0.359 |
| 8 | 0.325 | 0.336 | 0.307 | 0.318 | 0.301 | 0.308 | 0.298 | 0.282 | 0.281 | 0.289 | 0.265 | 0.296 | 0.268 | 0.310 |
| 9 | 0.286 | 0.295 | 0.269 | 0.279 | 0.264 | 0.270 | 0.262 | 0.247 | 0.246 | 0.254 | 0.232 | 0.260 | 0.235 | 0.272 |
| 10 | 0.254 | 0.263 | 0.240 | 0.249 | 0.235 | 0.241 | 0.233 | 0.220 | 0.219 | 0.226 | 0.207 | 0.231 | 0.209 | 0.242 |
| 11 | 0.229 | 0.237 | 0.216 | 0.224 | 0.211 | 0.216 | 0.209 | 0.198 | 0.197 | 0.203 | 0.186 | 0.207 | 0.188 | 0.218 |
| 12 | 0.208 | 0.215 | 0.196 | 0.203 | 0.192 | 0.196 | 0.190 | 0.179 | 0.179 | 0.184 | 0.168 | 0.188 | 0.170 | 0.198 |
| 13 | 0.191 | 0.197 | 0.179 | 0.186 | 0.176 | 0.180 | 0.174 | 0.164 | 0.163 | 0.168 | 0.154 | 0.172 | 0.155 | 0.181 |
| 14 | 0.176 | 0.182 | 0.165 | 0.171 | 0.162 | 0.165 | 0.160 | 0.151 | 0.150 | 0.155 | 0.142 | 0.158 | 0.143 | 0.167 |
| 15 | 0.163 | 0.168 | 0.153 | 0.158 | 0.150 | 0.153 | 0.148 | 0.140 | 0.139 | 0.143 | 0.131 | 0.146 | 0.132 | 0.155 |
| 16 | 0.152 | 0.157 | 0.142 | 0.147 | 0.139 | 0.143 | 0.138 | 0.130 | 0.129 | 0.133 | 0.122 | 0.136 | 0.123 | 0.144 |
| 17 | 0.142 | 0.147 | 0.133 | 0.138 | 0.130 | 0.133 | 0.129 | 0.121 | 0.121 | 0.125 | 0.114 | 0.127 | 0.115 | 0.135 |
| 18 | 0.133 | 0.138 | 0.125 | 0.129 | 0.122 | 0.125 | 0.121 | 0.114 | 0.113 | 0.117 | 0.107 | 0.119 | 0.107 | 0.126 |

## B. Entropy Rate Approximation

In order to attempt an entropy rate estimation, (3) is used, but considering words as symbols instead of characters. Table XI shows the values from $F_{1w}$ to $F_{5w}$ for texts EFE and LW1 to LW12.

TABLE XI
ENTROPY $Fnw$ (BITS/WORD)

| $Fnw$ | EFE | LW1 | LW2 | LW3 | LW4 | LW5 | LW6 | LW7 | LW8 | LW9 | LW10 | LW11 | LW12 |
|---|---|---|---|---|---|---|---|---|---|---|---|---|---|
| $F_{1w}$ | 10.43 | 9.91 | 9.80 | 9.71 | 9.86 | 9.97 | 9.70 | 9.69 | 10.19 | 10.00 | 10.32 | 9.51 | 9.77 |
| $F_{2w}$ | 4.79 | 5.00 | 4.56 | 4.55 | 4.24 | 4.25 | 4.27 | 3.92 | 3.73 | 3.85 | 3.50 | 3.92 | 3.33 |
| $F_{3w}$ | 0.97 | 1.07 | 0.88 | 0.65 | 0.61 | 0.52 | 0.60 | 0.51 | 0.14 | 0.19 | 0.15 | 0.30 | 0.03 |
| $F_{4w}$ | -0.16 | -0.19 | -0.21 | -0.30 | -0.27 | -0.31 | -0.28 | -0.24 | -0.37 | -0.36 | -0.38 | -0.34 | -0.37 |
| $F_{5w}$ | -0.28 | -0.29 | -0.28 | -0.31 | -0.30 | -0.31 | -0.31 | -0.30 | -0.32 | -0.32 | -0.31 | -0.32 | -0.32 |

For all texts analyzed in this work, after $F_{3w}$ the values of $F_{nw}$ become negative. Considering then that the limit in (4) is given by $F_{3w}$, an approximated value for the entropy rate, $H_L$, in bits/character, using (8), would be $H_{3\text{-word}}/\{3(\alpha+1)\}$ as is shown in Table XII.

TABLE XII
APPROXIMATE VALUES OF $H_L$ (BIT/LETTER)

| Text | $H_L$ |
|------|-------|
| EFE  | 0.93  |
| LW1  | 0.97  |
| LW2  | 0.89  |
| LW3  | 0.93  |
| LW4  | 0.88  |
| LW5  | 0.90  |
| LW6  | 0.87  |
| LW7  | 0.82  |
| LW8  | 0.83  |
| LW9  | 0.85  |
| LW10 | 0.78  |
| LW11 | 0.87  |
| LW12 | 0.80  |

In Table XII, the weighted average of $H_L$ is 0.90 bits/character. Since the weighted average of the number of different characters used in all of the natural Spanish text analyzed in this paper is 91.67, then the redundancy, $R$, would be approximately:

$$R = 1 - \frac{0.90}{\log_2(91.67)} \approx 86\%$$

The values in Table XII are a just an estimation since by definition, a sample of text of infinite length would be required for finding $H_L$ accurately.

AT1, AT2, AT3 were found to have essentially the same value of $H$ as for LW1, LW2, and LW3 for four-word symbols and beyond.

*C. The Spanish language constant*

Smoothing the 1-word curve in Fig. 6, the probability of the $r^{\text{th}}$ most frequent 1-word symbol is quite near to $0.08/r$, assuming $r$ is not too large. This constant relation is not observed for 2-word and 3-word symbols.

V. CONCLUSION

It is not an easy goal to calculate the entropy of a language with high precision due the dependency on the context of the text being analyzed. However some tendencies about entropy can be identified by analyzing reasonably long samples of text.

For all the text samples of natural Spanish analyzed in this work, $H_{max}$ was found to occur at 3-word symbols. For most of the literary works analyzed in this work, $n = 9$ was the value after which the $n$-word sequences become equiprobable, i.e., no $n$-word symbol repeats more than once on the string of text that every work represents. This observation however was not found to be true for the language of the news due to text reuse of partial news.

If 0.90 bit/letter can be regarded as a reasonable value for the entropy rate of Spanish, then the redundancy of the Spanish language should be around 86%.

An approximate value for the probability of the $r^{\text{th}}$ most frequent word in Spanish is $0.08/r$. Compared to the approximation for English ($0.1/r$) this means that, in general, in Spanish more words are used to convey the same meaning because the probability of words in Spanish is more spread among words than in English.

The log-log plots of Spanish language for both 2-word and 3-word symbols were found to have approximately constant but different negative slopes.

Finally, the artificial text produced using a first-order source model was found to approximate well in its statistical properties to natural Spanish for character, digram, trigram, and 1-word analysis. This corroborates the assertion that a stochastic model can be a good mathematical model for analyzing a discrete source, as was proposed by C. E. Shannon in 1948, although he was considering, of course, only the probabilistic side of the problem.

ACKNOWLEDGMENT

The author would like to thank Ms. Ana Mengotti, edition board director of the EFE press agency in Bogota (Colombia), for the news archive provided for this research.

**Fabio G. Guerrero** (M'93) received a B.Eng. degree in telecommunications engineering from the Universidad del Cauca, Popayan, Colombia, in 1992 and the M.Sc. degree in real-time electronic systems from Bradford University, U.K., in 1995. Currently, he works as Assistant Lecturer in the Department of Electrical and Electronics Engineering of the Universidad del Valle, Cali, Colombia, where he is also member of the Research Group SISTEL-UV on telecommunication systems. His research interests include digital communications, telecommunication systems modeling, and next-generation networks.